\newif\ifpeerreview

\ifpeerreview
\documentclass[peerreviewca]{IEEEtran}
\else
\documentclass[conference]{IEEEtran}
\IEEEoverridecommandlockouts
\fi

\usepackage[english]{babel}
\usepackage{amsmath}
\usepackage{graphicx}
\usepackage{textcomp}
\usepackage{xcolor}
\usepackage{tikz} 
\usepackage{pgfplots} 
\usetikzlibrary{pgfplots.groupplots}
\usepackage{float}
\usepackage{longtable}
\usepackage{multirow}
\usepackage{algorithm}
\usepackage{algpseudocodex}
\usepackage{booktabs}
\usepackage[marginpar]{todo}
\usepackage{balance}
\usepackage{soul}		
\usepackage{dblfloatfix}
\usepackage{subcaption}
\usepackage{nameref}
\usepackage{notoccite}
\usepackage{hyperref}
\usepackage{paralist}
\usetikzlibrary{patterns}

\def\BibTeX{{\rm B\kern-.05em{\sc i\kern-.025em b}\kern-.08em
    T\kern-.1667em\lower.7ex\hbox{E}\kern-.125emX}}
    
\graphicspath{{./img/}}

\pgfplotsset{compat = 1.8}
\usepgfplotslibrary{statistics}

\definecolor{my_green}{RGB}{0,158,115}
\definecolor{my_blue}{RGB}{86,180,233}
\definecolor{my_orange}{RGB}{230,159,0}
\definecolor{my_yellow}{RGB}{240,228,66}
\definecolor{my_red}{RGB}{213,94,0}
\definecolor{my_purple}{RGB}{204,121,167}

\pgfplotscreateplotcyclelist{ColorBlindFriendlyCycleList}{%
	solid, my_green, every mark/.append style={solid, fill=my_green}, mark=x\\%
	solid, my_blue, every mark/.append style={solid, fill=my_blue}, mark=*\\%
	solid, my_orange, every mark/.append style={solid, fill=my_orange}, mark=otimes*\\%
	solid, my_yellow, every mark/.append style={solid, fill=my_yellow}, mark=triangle*\\%
	solid, my_red, every mark/.append style={solid, fill=my_red},mark=diamond*\\%
	solid, my_purple, every mark/.append style={solid, fill=my_purple},mark=*\\%
	dashed, my_green, every mark/.append style={solid, fill=my_green},mark=square*\\%
	dashed, my_blue, every mark/.append style={solid, fill=my_blue},mark=otimes*\\%
}

\pgfplotscreateplotcyclelist{ColorBlindFriendlyCycleListBar}{%
	solid, fill=my_green, every mark/.append style={solid, fill=my_green}\\%
	solid, fill=my_blue, every mark/.append style={solid, fill=my_blue}\\%
	solid, fill=my_orange, every mark/.append style={solid, fill=my_orange}\\%
	solid, fill=my_yellow, every mark/.append style={solid, fill=my_yellow}\\%
	solid, fill=my_red, every mark/.append style={solid, fill=my_red}\\%
	solid, fill=my_purple, every mark/.append style={solid, fill=my_purple}\\%
	solid, fill=black, every mark/.append style={solid, fill=black}\\%
	solid, fill=white, every mark/.append style={solid, fill=white}\\%
}

\pgfplotscreateplotcyclelist{ColorBlindFriendlyCycleListBarPattern}{%
	solid, fill=my_green, every mark/.append style={solid, fill=my_green}, postaction={pattern=north east lines}\\%
	solid, fill=my_blue, every mark/.append style={solid, fill=my_blue}, postaction={pattern=north east lines}\\%
	solid, fill=my_orange, every mark/.append style={solid, fill=my_orange}, postaction={pattern=north east lines}\\%
	solid, fill=my_yellow, every mark/.append style={solid, fill=my_yellow}, postaction={pattern=north east lines}\\%
	solid, fill=my_red, every mark/.append style={solid, fill=my_red}, postaction={pattern=north east lines}\\%
	solid, fill=my_purple, every mark/.append style={solid, fill=my_purple}, postaction={pattern=north east lines}\\%
	solid, fill=black, every mark/.append style={solid, fill=black}, postaction={pattern=north east lines}\\%
	solid, fill=white, every mark/.append style={solid, fill=white}, postaction={pattern=north east lines}\\%
}

\pgfplotsset{
    discard if not/.style 2 args={
        x filter/.code={
            \edef\tempa{\thisrow{#1}}
            \edef\tempb{#2}
            \ifx\tempa\tempb
            \else
                
            \fi
        }
    }
}

\pgfplotsset{
    legend image with text/.style={
        legend image code/.code={%
            \node[anchor=center] at (0.3cm,0cm) {#1};
        }
    },
}

\algblock{ParFor}{EndParFor}
\algnewcommand\algorithmicparfor{\textbf{parfor}}
\algnewcommand\algorithmicpardo{\textbf{do}}
\algnewcommand\algorithmicendparfor{\textbf{end\ parfor}}
\algrenewtext{ParFor}[1]{\algorithmicparfor\ #1\ \algorithmicpardo}
\algrenewtext{EndParFor}{\algorithmicendparfor}

\usepackage{environ}
\NewEnviron{scaled_IEEEeqnarray}{%
	\begin{IEEEeqnarray}{l}
		\scalebox{0.8}{$\BODY$}
	\end{IEEEeqnarray}
}

\newcommand{\papertitle}{Tensor Program Optimization for the RISC-V Vector Extension Using Probabilistic Programs}
\newcommand{\fundingagency}{German Federal Ministry of Education and Research}
\newcommand{\fundingproject}{"GreenEdge-FuE"}
\newcommand{\fundingnumber}{16ME0517K}

\begin{document}

\ifpeerreview
\title{\papertitle}
\else
\title{\papertitle \thanks{This research was funded by the \fundingagency\: within the projects \fundingproject (funding nr. \fundingnumber) and the CHIPS JU project "ISOLDE" (project nr. 101112274, BMFTR funding nr. 16MEE0334).}}
\fi
	
\author{
	\IEEEauthorblockN{Federico Nicolás Peccia, Frederik Haxel, Oliver Bringmann}
	\IEEEauthorblockA{FZI Research Center for Information Technology, University of Tübingen\\
		Germany \\
		peccia@fzi.de,haxel@fzi.de,oliver.bringman@uni-tuebingen.de}
}

\ifpeerreview
\IEEEpeerreviewmaketitle 
\else
\maketitle
\fi

\begin{abstract}





RISC-V provides a flexible and scalable platform for applications ranging from embedded devices to high-performance computing clusters. Particularly, its RISC-V Vector Extension (RVV) becomes of interest for the acceleration of AI workloads. But writing software that efficiently utilizes the vector units of RISC-V CPUs without expert knowledge requires the programmer to rely on the autovectorization features of compilers or hand-crafted libraries like muRISCV-NN. Smarter approaches, like autotuning frameworks, have been missing the integration with the RISC-V RVV extension, thus heavily limiting the efficient deployment of complex AI workloads. In this paper, we present a workflow based on the TVM compiler to efficiently map AI workloads onto RISC-V vector units. Instead of relying on hand-crafted libraries, we integrated the RVV extension into TVM's MetaSchedule framework, a probabilistic program framework for tensor operation tuning. We implemented different RISC-V SoCs on an FPGA and tuned a wide range of AI workloads on them. We found that our proposal shows a mean improvement of 46\% in execution latency when compared against the autovectorization feature of GCC, and 29\% against muRISCV-NN. Moreover, the binary resulting from our proposal has a smaller code memory footprint, making it more suitable for embedded devices. Finally, we also evaluated our solution on a commercially available RISC-V SoC implementing the RVV 1.0 Vector Extension and found our solution is able to find mappings that are 35\% faster on average than the ones proposed by LLVM. We open-sourced our proposal for the community to expand it to target other RISC-V extensions.

\end{abstract}

\begin{IEEEkeywords}
    RISC-V, RVV, TVM, Vector processors
\end{IEEEkeywords}

\section{Introduction}
\label{section:introduction}

The execution of AI models is nowadays a task that permeates all computing levels, from High Performance Computing (HPC) servers to embedded devices. Given its open-source nature, scalability, and broadly ratified extensions, the RISC-V ISA presents itself as an ideal candidate to accelerate the execution of these AI workloads across this wide range of hardware platforms.

Since its Vector Extension (RVV) has been ratified, numerous commercial \cite{9138983,bpi} and research platforms \cite{zhao2024instructionschedulingsaturnvector,Spatz2023, 9912071} have added support for it. But although the hardware is now available, the software support to deploy AI workloads that efficiently use the vector unit is still lacking. Even though compilers like GCC and LLVM provide autovectorization features, these do not always use the vector unit as efficiently as manually programmed kernels \cite{9802745}.

Another option is to use libraries of hand-crafted kernels like muRISCV-NN \cite{muriscv-nn}, which have been demonstrated to provide higher speedups than the compiler's autovectorization. But these libraries do not adapt to changes in the hardware: for example, there is no guarantee that a kernel written for a RISC-V CPU with a 1 Mb L2 cache will still be the best option for one with a bigger L2 cache, where the scheduling of the AI workload could benefit from different data reuse. This gets even more complicated given that vector units from different vendors can expose different performances because of differences in their microarchitecture implementation. This is where tuning the AI workload for the particular target hardware using frameworks like AutoTVM \cite{autotvm} or MetaSchedule \cite{shao2022tensorprogramoptimizationprobabilistic} becomes advantageous. But so far, the RISC-V RVV extension has been missing from these kinds of frameworks, heavily limiting the deployment of AI workloads on RISC-V vector units.

In this paper, we extended the MetaSchedule framework of TVM with \textit{tensor intrinsics} that make use of the RISC-V RVV extension. These intrinsic, together with the probabilistic sampling of the possible mappings provided by MetaSchedule enable an efficient exploration of the design space of possible schedule candidates of each tensor operation onto the RISC-V vector unit.

To evaluate our proposal, instead of relying on simulators, we perform a broad study implementing different versions of a RISC-V System-on-Chip (SoC), together with a vector unit that supports the RVV 1.0 extension, on an AMD ZCU102 FPGA. We tuned multiple AI workloads on each hardware version and compared our work against muRISCV-NN and the autovectorization feature from GCC 14. We found that, for all the evaluated hardware configurations, our proposal outperforms both of them. We also analyzed instruction traces to verify that our schedules utilize the vector register file more efficiently than muRISCV-NN. We even evaluated vector units with different hardware parameters and confirmed that our solution still finds better schedules than the other approaches for each hardware version. Then, to demonstrate that our integration is also able to target commercially available boards, we tuned several AI workloads on the Banana Pi BPI-F3 board, which provides an octa-core RISC-V SoC with 256-bit RVV 1.0 compatible vector units. We used the full capabilities of the TVM runtime to execute these and found that our tuned workloads are better than the ones executed using plain TVM (enabling the autovectorization of LLVM 19).


This work demonstrates the advantages of probabilistic program exploration for the acceleration of AI workloads using the RISCV RVV extension. The open-source nature of our proposal will allow other works to expand this approach using other RISCV extensions, for example, Packed SIMD. In addition, our integration is able to target both the microTVM runtime (for embedded devices running bare metal or an RTOS) and the full TVM runtime (for more capable embedded devices or servers). This will enable the deployment of AI workloads on a wide range of RISC-V platforms implementing the RVV extension.

The rest of this paper is organized as follows. First, Section \ref{section:related_work} presents the current options for developers to map AI workloads onto the vector units of RISC-V CPUs. Then, Section \ref{section:proposal} presents our extension to the MetaSchedule framework to enable the exploration of different mapping possibilities of these AI workloads using the RISC-V RVV extension. In Section \ref{section:eval}, we demonstrate that the mappings found by our proposal surpass the ones from previous works on a variety of hardware configurations. Finally, Section \ref{section:conclusions} concludes the work and discusses future research directions.

\section{Related work}
\label{section:related_work}

    The RISC-V RVV 1.0 Vector Extension introduces a flexible vector processing model in order to support a wide range of applications, from digital signal processing (DSP), graphics or AI workloads, and targets platforms ranging from embedded systems to high-performance computing (HPC). It provides 32 vector registers, each up to VLEN bits wide. Using a combination of SEW (Selected Element Width, the actual width of each element in the vector to be processed) and LMUL (Vector Register Group Multiplier, which allows the programmer to group multiple vector registers together to process longer sequences of elements) the program can define, during runtime, how many elements each vector instructions is actually going to operate on (VL, or Vector Length).

    In order to accelerate applications using this RISC-V extension, programmers can always write their program and insert the specific assembly instructions to offload computation to the vector unit. But this is cumbersome and requires a lot of expertise. This is why GCC provides C intrinsics to abstract the programmer from the actual RVV instructions call. But still, writing its own code that uses vector intrinsics requires a lot of effort, even more the more complex the application gets. To automate this process, compilers like GCC 14 and LLVM 19 already provide autovectorization features to automatically offload operations to the vector unit. However, these are still heavily dependent on the way the code is written.
    
    muRISCV-NN \cite{muriscv-nn} provided a compilation flow to accelerate AI workloads using the RVV extension for 32-bit RISC-V CPUs. They used the existing CMSIS-NN integration from TVM to generate C code, and then replaced the calls to the ARM-based CMSIS-NN kernels with their own hand-crafted RISC-V RVV kernels. They evaluated their generated C code on multiple simulators to demonstrate the speedup in comparison with the autovectorization features of the compilers.

    MetaSchedule \cite{shao2022tensorprogramoptimizationprobabilistic} is part of the TVM Deep Learning compiler and enables the exploration of possible mappings of tensor operations onto a particular hardware using probabilistic programs. During the tuning process, MetaSchedule: \begin{inparaenum}
        \item generates mapping candidates based on the sampling of probabilistic schedule transformations,
        \item evaluates the proposed candidate on the hardware and
        \item uses its performance to tune a cost model that guides the next candidates' selection process
    \end{inparaenum}. In the end, the candidate with the lower latency is returned. \textit{Tensor intrinsics} mapping a certain minimal tensor operation onto target-specific instructions or library calls need to be defined in order for each candidate to use the features provided by the target hardware. Much work has been done to define these \textit{tensor intrinsics} to map operations onto ARM CPUs using the NEON instructions or onto NVIDIA GPUs using CUDA. But until now, the RISC-V RVV extension has been missing, forcing developers to rely on the auto-vectorization capabilities of the compiler or hand-crafted libraries like muRISCV-NN.


\section{Proposal}
\label{section:proposal}

    \begin{figure}[!t]
		\centering
		\includegraphics[width=0.48\textwidth]{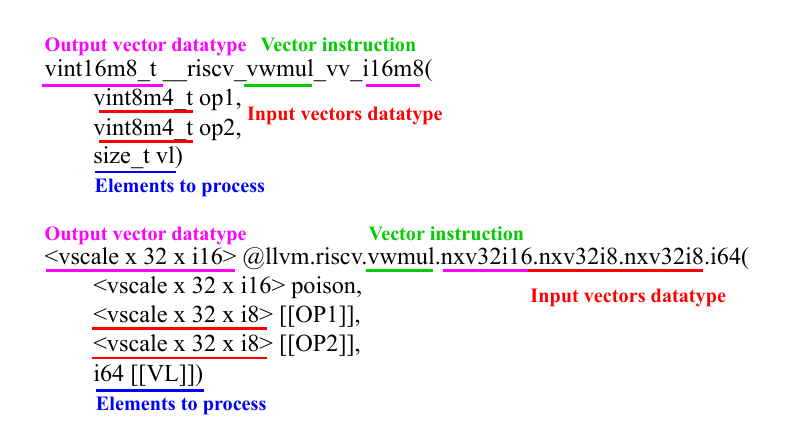}
		\caption{For the standard vector-vector elementwise multiplication with widening instruction, we show how to parameterize the GCC (top) and LLVM (bottom) intrinsics based on the datatype of the input and output vectors.}
		\label{fig:intrinsics_datatype}
	\end{figure}
    
    We propose to integrate the RISC-V RVV extension into the TVM Deep Learning compiler and take advantage of its MetaSchedule autotuning framework to find efficient mappings of AI tensor operations onto the vector unit. As explained in Section \ref{section:related_work}, MetaSchedule works with \textit{tensor intrinsics}. These are composed of two parts: a \textit{definition} of a small tensor operation that can be accelerated using instructions available in the target hardware, and an \textit{implementation}, where the calls to the actual hardware interfaces are used to execute the operation described in the \textit{definition}. The tensor operations of the model being optimized are then tiled to generate smaller operations that are then matched to the available \textit{tensor intrinsics}, and the sections matching the available \textit{definitions} are replaced with the appropriate \textit{implementations}.

    Implementing these \textit{tensor intrinsics} allows us to then use TVM to target different hardware platforms. For example, we can use its C code generation capabilities and its microTVM runtime to execute the best candidate schedules found by MetaSchedule on baremetal or RTOS-based systems. Or we can generate a shared library using TVM's LLVM integration to target more powerful devices able to run the entire TVM runtime. This requires us to define two different \textit{tensor intrinsics}. When targeting microTVM, our \textit{implementations} generate calls to the GCC RVV intrinsics to interface with the vector unit. For LLVM-based targets, we used the LLVM RVV intrinsics. As both the GCC and the LLVM RVV intrinsics are easily parameterized in terms of the datatypes of the vectors involved in the operation (Figure \ref{fig:intrinsics_datatype}), we can define generic \textit{implementations} that work for any datatype, and then generate the appropriate calls based on the datatypes of the inputs and output of the \textit{definition}. This allows us to target not only \textit{int8} tensor operations (like muRISCV-NN) but also \textit{float16} and \textit{float32} ones.


    \begin{figure}[!t]
		\centering
		\includegraphics[width=0.48\textwidth]{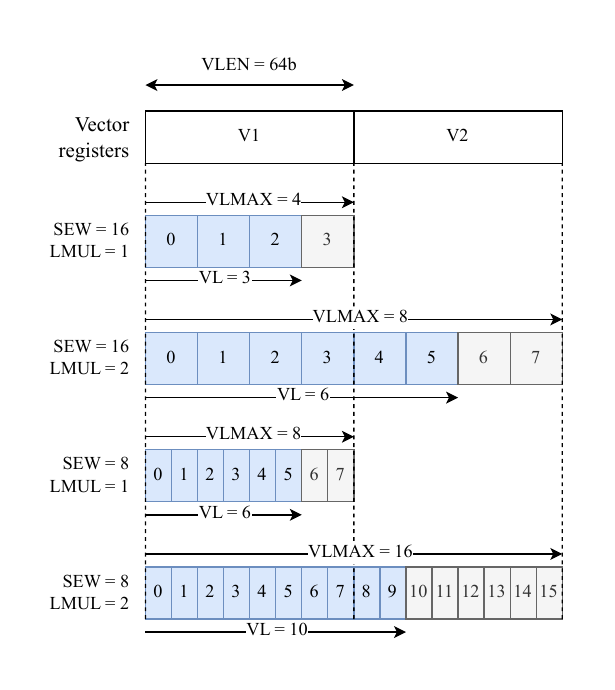}
		\caption{Relationship between hardware (VLEN) and software (SEW, LMUL, VL) parameters in the RVV ISA extension.}
		\label{fig:rvv_hw_sw}
	\end{figure}
    
    One challenge of implementing tensor operations using the RISC-V RVV extension comes from the flexibility of this ISA extension. Not only do the hardware parameters of the vector unit impact our \textit{tensor intrinsics}, but we also need to select the appropriate software parameters to be configured during runtime. Figure \ref{fig:rvv_hw_sw} shows the relation between these parameters. VLEN defines the length in bits of each vector register, and as such it is fixed by the hardware. The Selected Element Width (SEW) configures during runtime the width of each element to be processed, and thus it is defined by the datatype of the tensor operation. The Vector Register Group Multiplier (LMUL) is also a programmable parameter, which defines how many vector registers are going to be used together for a vector instruction. Together, SEW and LMUL define the maximum amount of elements that can be processed, and is calculated as presented in Equation \eqref{eq:vlmax}. Finally, VL determines the \textit{actual} amount of elements that are going to be processed by one particular vector instruction. In order to be able to process as many elements as possible, we decided to use LMUL = 8 (the maximum value) in our \textit{tensor intrinsics}.
    

    \begin{equation}
        VLMAX [elements] = \frac{VLEN [bits] * LMUL}{SEW [bits]}
        \label{eq:vlmax}
    \end{equation}

    Algorithms \ref{fig:rvv_multivmul} and \ref{fig:rvv_vmacc} present the pseudocode for the \textit{implementation} of the \textit{tensor intrinsics} we propose to integrate into TVM. Algorithm \ref{fig:rvv_multivmul} represents a typical operation found in AI workloads, mostly in fully connected, convolution, or attention layers. The intrinsic in Algorithm \ref{fig:rvv_vmacc} is provided to efficiently map layers that do not require a reduction operation (for example, depthwise convolutions). Although these intrinsics seem simple in their structure, it is their integration into the probabilistic program tuning framework MetaSchedule which enables an efficient deployment of workloads on RISC-V vector units.

    \begin{algorithm}[!t]
        \caption{Pseudocode for the intrinsic for vector-matrix multiplication.}
        \label{fig:rvv_multivmul}
		\small
		\begin{algorithmic}[1]
            \Require VL $\leq$ VLMAX
			\Function{rvv\_multivmul}{A[VL],B[J,VL],C[J]}
                \LComment{Load entire A vector onto a vector register, and C vector for accumulation later on}
                \State A\_vec = \textit{vle}(\&A, VL)
                \State C\_vec = \textit{vle}(\&C, J)
                \For{$j = 0, \dots, $J$-1$}
                    \LComment{Prepare vector register for sumation reduction}
                    \State red\_vec = \textit{vmv}(0, 1)
                    \LComment{Load one row of B matrix}
                    \State B\_vec = \textit{vle}(\&B[$j$][0], VL)
                    \LComment{Element-wise multiplication}
                    \State mult\_vec = \textit{vmul}(A\_vec, B\_vec, VL)
                    \LComment{Sum all individual multiplication results}
                    \State red\_vec = \textit{vredsum}(mult\_vec, red\_vec, VL)
                    \LComment{Merge reduced result onto final output register}
                    \If{$j == 0$}
                        \State out\_vec = \textit{vmv}(red\_vec, 1)
                    \Else
                        \State out\_vec = \textit{vslideup}(out\_vec, \textit{vmv}(red\_vec, 1), $j$, $j+1$)
                    \EndIf
                \EndFor
                \LComment{Accumulate multiplication results with previous C values}
                \State out\_vec = \textit{vadd}(out\_vec, C\_vec, J)
                \LComment{Store result back onto the C vector}
                \State \textit{vse}(\&C, out\_vec, J)
                \State \Return
            \EndFunction
		\end{algorithmic}
	\end{algorithm}
    
    \begin{algorithm}[!t]
        \caption{Pseudocode for the intrinsic for vector-vector elementwise multiplication with accumulation.}
        \label{fig:rvv_vmacc}
		\small
		\begin{algorithmic}[1]
            \Require VL $\leq$ VLMAX
			\Function{rvv\_vmacc}{A[VL],B[VL],C[VL]}
                \LComment{Load entire A, B and C vectors onto vector registers}
                \State A\_vec = \textit{vle}(\&A, VL)
                \State B\_vec = \textit{vle}(\&B, VL)
                \State out\_vec = \textit{vle}(\&C, VL)
                \LComment{Perform multiplication, accumulating with out\_vec}
                \State out\_vec = \textit{vmacc}(out\_vec, A\_vec, B\_vec, VL)
                \LComment{Store result back onto the C vector}
                \State \textit{vse}(\&C, out\_vec, VL)
                \State \Return
            \EndFunction
		\end{algorithmic}
	\end{algorithm}
    
    \begin{figure*}[!t]
			\begin{tikzpicture}
			\begin{axis}[
                    title=Int8,
                    ymin=0.1,
                    ymax=1000000,
                    width=0.45\textwidth,
                    height=0.15\textheight,
                    ylabel={Latency [ms]},
                    ybar,
                    ymode=log,
                    log origin=infty,
                    bar width=0.12cm,
                    enlarge y limits=false,
                    legend columns=4,
                    legend style={at={(0.01,0.98)},anchor=north west, font=\tiny},
                    legend cell align={left},
                    x label style={font=\footnotesize},
                    x tick label style={rotate=90, font=\footnotesize},
                    y tick label style={font=\footnotesize},
					ytick pos=left,
                    symbolic x coords={
                        16,
                        32,
                        64,
                        128,
                        256,
                        512,
                    },
                    xtick=data,
                    xticklabel=\empty,
                    ytick={1,100,10000,1000000, 100000000},
				    cycle list name=ColorBlindFriendlyCycleListBar,
				]
                \addplot+
                        plot table[discard if not={type}{non-tuned}, x=matrix_size,y={latency [ms]},col sep=comma] {./data/saturn_tests_1024_int8.csv};
                \addlegendentry{Non tuned}
                \addplot+
                        plot table[discard if not={type}{non-tuned-O3}, x=matrix_size,y={latency [ms]},col sep=comma] {./data/saturn_tests_1024_int8.csv};
                \addlegendentry{Non tuned (-O3)}
                \addplot+
                        plot table[discard if not={type}{muRISCV-NN}, x=matrix_size,y={latency [ms]},col sep=comma] {./data/saturn_tests_1024_int8.csv};
                \addlegendentry{muRISCV-NN}
                \addplot+
                        plot table[discard if not={type}{tuned-RVV}, x=matrix_size,y={latency [ms]},col sep=comma] {./data/saturn_tests_1024_int8.csv};
                \addlegendentry{Ours}
			\end{axis}
		\end{tikzpicture}
        \begin{tikzpicture}
			\begin{axis}[
                    title=Float16,
                    ymin=0.1,
                    ymax=1000000,
                    width=0.36\textwidth,
                    height=0.15\textheight,
                    ybar,
                    ymode=log,
                    log origin=infty,
                    bar width=0.12cm,
                    enlarge y limits=false,
                    legend columns=2,
                    legend style={at={(0.01,0.98)},anchor=north west, font=\footnotesize},
                    legend cell align={left},
                    x label style={font=\footnotesize},
                    x tick label style={rotate=90, font=\footnotesize},
                    y tick label style={font=\footnotesize},
					ytick pos=left,
                    symbolic x coords={
                        16,
                        32,
                        64,
                        128,
                        256,
                        512,
                    },
                    xtick=data,
                    ytick={1,100,10000,1000000, 100000000},
                    xticklabel=\empty,
                    yticklabel=\empty,
				    cycle list name=ColorBlindFriendlyCycleListBar,
				]
                \addplot+
                        plot table[discard if not={type}{non-tuned}, x=matrix_size,y={latency [ms]},col sep=comma] {./data/saturn_tests_1024_float16.csv};
                \addplot+
                        plot table[discard if not={type}{non-tuned-O3}, x=matrix_size,y={latency [ms]},col sep=comma] {./data/saturn_tests_1024_float16.csv};
                \pgfplotsset{cycle list shift=1}
                \addplot+
                        plot table[discard if not={type}{tuned-RVV}, x=matrix_size,y={latency [ms]},col sep=comma] {./data/saturn_tests_1024_float16.csv};
			\end{axis}
		\end{tikzpicture}
        \begin{tikzpicture}
			\begin{axis}[
                    title=Float32,
                    ymin=0.1,
                    ymax=1000000,
                    width=0.36\textwidth,
                    height=0.15\textheight,
                    ybar,
                    ymode=log,
                    log origin=infty,
                    bar width=0.12cm,
                    enlarge y limits=false,
                    legend columns=2,
                    legend style={at={(0.01,0.98)},anchor=north west, font=\footnotesize},
                    legend cell align={left},
                    x label style={font=\footnotesize},
                    x tick label style={rotate=90, font=\footnotesize},
                    y tick label style={font=\footnotesize},
					ytick pos=left,
                    symbolic x coords={
                        16,
                        32,
                        64,
                        128,
                        256,
                        512,
                    },
                    xtick=data,
                    ytick={1,100,10000,1000000, 100000000},
                    yticklabel=\empty,
                    xticklabel=\empty,
				    cycle list name=ColorBlindFriendlyCycleListBar,
				]
                \addplot+
                        plot table[discard if not={type}{non-tuned}, x=matrix_size,y={latency [ms]},col sep=comma] {./data/saturn_tests_1024_float32.csv};
                \addplot+
                        plot table[discard if not={type}{non-tuned-O3}, x=matrix_size,y={latency [ms]},col sep=comma] {./data/saturn_tests_1024_float32.csv};
                \pgfplotsset{cycle list shift=1}
                \addplot+
                        plot table[discard if not={type}{tuned-RVV}, x=matrix_size,y={latency [ms]},col sep=comma] {./data/saturn_tests_1024_float32.csv};
			\end{axis}
		\end{tikzpicture}
        \hspace*{\fill}

        \vspace{-0.5cm}
        \begin{tikzpicture}
			\begin{axis}[
                    ymin=0,
                    ymax=100,
                    width=0.45\textwidth,
                    height=0.15\textheight,
                    xlabel={Matrices dimensions},
                    ylabel={Speedup [\%]},
                    ybar,
                    bar width=0.12cm,
                    enlarge y limits=false,
                    legend columns=2,
                    legend style={at={(0.01,0.9)},anchor=north west, font=\footnotesize},
                    x label style={font=\footnotesize},
                    x tick label style={font=\footnotesize},
                    y tick label style={font=\footnotesize},
					ytick pos=left,
                    symbolic x coords={
                        16,
                        32,
                        64,
                        128,
                        256,
                        512,
                    },
                    xtick=data,
				    cycle list name=ColorBlindFriendlyCycleListBar,
				]
                \addplot+ plot coordinates {(16,0) (32,0) (64,0) (128,0) (256,0) (512,0)};
                \addplot+
                        plot table[discard if not={type}{non-tuned-O3}, x=matrix_size,y=perc,col sep=comma] {./data/saturn_tests_1024_int8.csv};
                \addplot+
                        plot table[discard if not={type}{muRISCV-NN}, x=matrix_size,y=perc,col sep=comma] {./data/saturn_tests_1024_int8.csv};
                \addplot+
                        plot table[discard if not={type}{tuned-RVV}, x=matrix_size,y=perc,col sep=comma] {./data/saturn_tests_1024_int8.csv};
			\end{axis}
		\end{tikzpicture}
        \begin{tikzpicture}
			\begin{axis}[
                    ymin=0,
                    ymax=100,
                    width=0.36\textwidth,
                    height=0.15\textheight,
                    xlabel={Matrices dimensions},
                    ybar,
                    bar width=0.12cm,
                    enlarge y limits=false,
                    legend columns=2,
                    legend style={at={(0.01,0.9)},anchor=north west, font=\footnotesize},
                    x label style={font=\footnotesize},
                    x tick label style={font=\footnotesize},
                    y tick label style={font=\footnotesize},
					ytick pos=left,
                    symbolic x coords={
                        16,
                        32,
                        64,
                        128,
                        256,
                        512,
                    },
                    xtick=data,
                    yticklabel=\empty,
				    cycle list name=ColorBlindFriendlyCycleListBar,
				]
                \addplot+ plot coordinates {(16,0) (32,0) (64,0) (128,0) (256,0) (512,0)};
                \addplot+
                        plot table[discard if not={type}{non-tuned-O3}, x=matrix_size,y=perc,col sep=comma] {./data/saturn_tests_1024_float16.csv};
                \pgfplotsset{cycle list shift=1}
                \addplot+
                        plot table[discard if not={type}{tuned-RVV}, x=matrix_size,y=perc,col sep=comma] {./data/saturn_tests_1024_float16.csv};
			\end{axis}
		\end{tikzpicture}
        \begin{tikzpicture}
			\begin{axis}[
                    ymin=0,
                    ymax=100,
                    width=0.36\textwidth,
                    height=0.15\textheight,
                    xlabel={Matrices dimensions},
                    ybar,
                    bar width=0.12cm,
                    enlarge y limits=false,
                    legend columns=2,
                    legend style={at={(0.01,0.9)},anchor=north west, font=\footnotesize},
                    x label style={font=\footnotesize},
                    x tick label style={font=\footnotesize},
                    y tick label style={font=\footnotesize},
					ytick pos=left,
                    symbolic x coords={
                        16,
                        32,
                        64,
                        128,
                        256,
                        512,
                    },
                    xtick=data,
                    yticklabel=\empty,
				    cycle list name=ColorBlindFriendlyCycleListBar,
				]
                \addplot+ plot coordinates {(16,0) (32,0) (64,0) (128,0) (256,0) (512,0)};
                \addplot+
                        plot table[discard if not={type}{non-tuned-O3}, x=matrix_size,y=perc,col sep=comma] {./data/saturn_tests_1024_float32.csv};
                \pgfplotsset{cycle list shift=1}
                \addplot+
                        plot table[discard if not={type}{tuned-RVV}, x=matrix_size,y=perc,col sep=comma] {./data/saturn_tests_1024_float32.csv};
			\end{axis}
		\end{tikzpicture}
        \hspace*{\fill}
        \caption{Benchmarking of matrix multiplications on the Saturn Vector Unit (VLEN=1024). Speedup is calculated using "Non tuned" as baseline.}
        \label{fig:saturn_1024_test}
    \end{figure*}
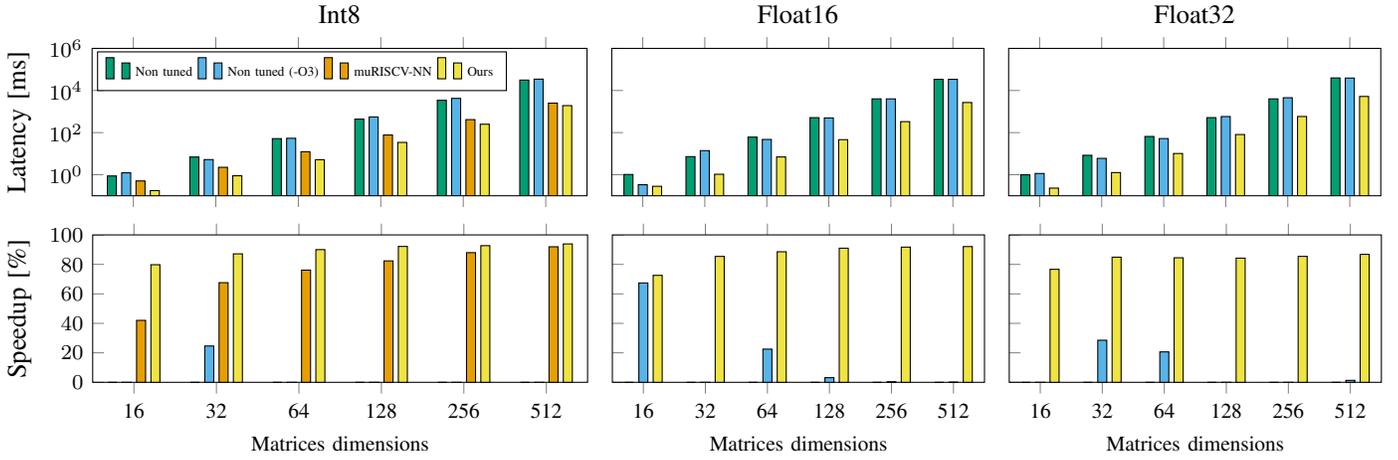

    Still, the ability of the RISC-V RVV extension to work with variable vector lengths presents an additional challenge. Indeed, LMUL and SEW define the \textit{maximum} possible elements that can be processed by one vector instruction, but then each intrinsic (GCC and LLVM) requires to pass VL as a parameter. This is a problem because MetaSchedule requires the shape of the input tensors of the \textit{definition} of the \textit{tensor intrinsic} to be static. So, if we configure a \textit{definition} with a VL much smaller than VLMAX, we lose the opportunity to accelerate more efficiently operations that are bigger than VL (but smaller than VLMAX). On the other hand, if we configure the \textit{definition} with VL = VLMAX, tensor operations with input shapes smaller than this value will not be matched by MetaSchedule and thus will not be accelerated using our \textit{implementation}.
    
    To solve this, we register into TVM multiple versions of the same \textit{tensor intrinsics}, starting with VL = VLMAX and then halving VL for each version, until VL = 4 (we found empirically that, if the shapes of the tensors are smaller than 4, using the vector unit does not provide a significant speedup). During the tuning process, MetaSchedule will then try to map each tensor operation using all these available intrinsics with different VLs, and select the one appropriate for it. In this way, bigger operations can take advantage of an intrinsic with VL = VLMAX, while smaller operations can still be accelerated by the vector unit using a smaller VL.

    For the intrinsic presented in Algorithm \ref{fig:rvv_multivmul}, we also need to select an appropriate J for MetaSchedule to be able to match the related tensor operation. As such, we selected J = VLEN/32. This allows us to accumulate all the results of the intermediate reductions into one vector register and then write the entire register content onto memory. However, to also be able to map matrix multiplications that have a smaller J (for very small workloads), we also register into MetaSchedule an intrinsic version with J = 1.

\section{Evaluation}
\label{section:eval} 
    
    %
    To validate the performance of the programs found by our proposal, we tune tensor programs on different platforms. We use a workflow similar to the Chipyard generator framework \cite{chipyard} to generate SoCs containing a Rocket CPU with a Saturn Vector Unit \cite{zhao2024instructionschedulingsaturnvector} (with different VLENs) and an L2 cache of 512 kB, and implement them on a ZCU102 FPGA board with a clock frequency of 100 MHz. On the other hand, we also use the commercially available Banana Pi BPI-F3 board \cite{bpi} (which has a VLEN of 256 bits, 2 MB L2 cache and an operating frequency of 1.6 GHz) as a target. In terms of runtime software, for the FPGA-based SoCs, we use TVM to generate C code and compile it together with the microTVM runtime and the Zephyr RTOS \cite{zephyr} using GCC 14. For the Banana Pi BPI-F3 board, we use Ubuntu 24.04 as operating system, and use TVM and LLVM 19 to generate a shared library, which is then executed on the board using the TVM runtime\footnote{Because of limitations in the existing TVM-LLVM codegen, we don't combine each intermediate output in a temporal register and then copy the entire vector to memory (lines 15 to 22 in Figure \ref{fig:rvv_multivmul}). Instead, we copy each intermediate output to memory as soon as it is generated.}.

    For each tensor program, we compare the following scenarios against our proposed vector intrinsics. For the FPGA-based experiments, we first execute the generated C code as it is, without vector instructions (\textit{Non tuned} version, compiled with GCC's \textbf{-Os} flag). We then compile the same code with the GCC \textbf{-O3} flag, which enables the autovectorization feature of the compiler (\textit{Non tuned (-O3)}). We also execute the tensor programs using the muRISCV-NN integration (\textit{muRISCV-NN}). For the Banana Pi based experiments, we first compile the generated programs using LLVM without enabling the vector instructions (\textit{Non tuned}). We then use the LLVM autovectorization feature (\textit{Non tuned (v)}).
    
    \subsection{Matrix multiplications}
    \label{section:eval_matrix}
    
    First, we evaluate the performance of matrix multiplications in the form $C^{m\times n}=A^{m\times k}\times B^{k\times n}+D^{m\times n}$ across multiple square sizes ($m=n=k$). For the \textit{float32} versions, all matrices are of type \textit{float32}. But for the \textit{int8} versions, we define the operation as it normally appears in Quantized Neural Networks \cite{DBLP:journals/corr/abs-1712-05877}. Matrices $A$ and $B$ are of type \textit{int8}. Their multiplication results in an $m\times n$ \textit{int32} matrix, which is added to the $D$ matrix, also of type \textit{int32}. Finally, the resulting matrix is then converted to \textit{int8} using a requantization operator.

    For the tuned versions, we executed the MetaSchedule procedure of TVM for 100 iterations. As explained in Section \ref{section:related_work}, for each iteration, MetaSchedule generates a new program candidate, compiles it for the target hardware, flashes the hardware, executes the program and reads its latency. When generating C code and compiling it together with Zephyr, this process takes between 9 to 12 seconds per iteration. This is a downside of this kind of autotuning processes, but still, measuring 100 candidates still finishes in a timely fashion (less than 20 minutes).
    
    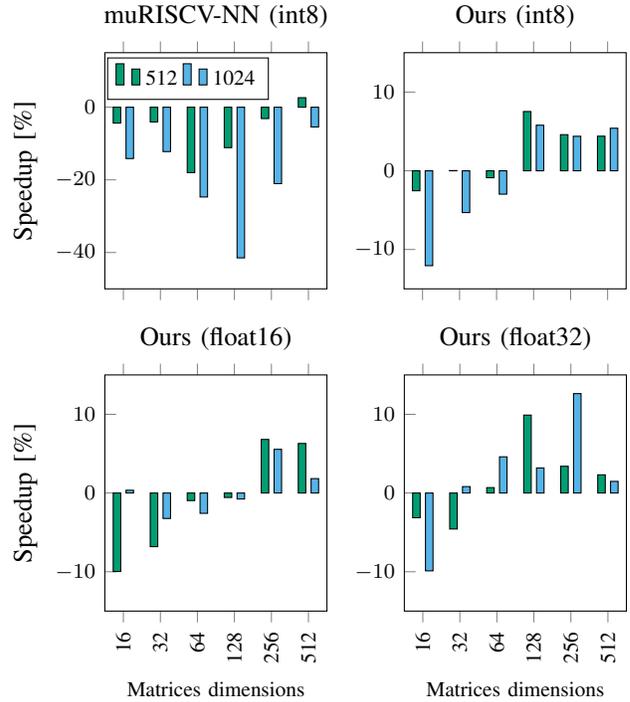
\begin{figure}[!t]
			\begin{tikzpicture}
			\begin{axis}[
                    title=muRISCV-NN (int8),
                    ymin=-50,
                    ymax=15,
                    width=0.25\textwidth,
                    height=0.2\textheight,
                    ylabel={Speedup [\%]},
                    ybar,
                    bar width=0.1cm,
                    enlarge y limits=false,
                    legend columns=2,
                    legend style={at={(0.01,0.98)},anchor=north west, font=\footnotesize},
                    legend cell align={left},
                    x label style={font=\footnotesize},
                    x tick label style={rotate=90, font=\footnotesize},
                    y tick label style={font=\footnotesize},
					ytick pos=left,
                    symbolic x coords={
                        16,
                        32,
                        64,
                        128,
                        256,
                        512,
                    },
                    xtick=data,
                    xticklabel=\empty,
				    cycle list name=ColorBlindFriendlyCycleListBar,
				]
                
                \addplot+
                        plot table[discard if not={VLEN}{512}, x=matrix_size,y={vlen_improvement},col sep=comma] {./data/saturn_tests_vlen_int8_muRISCV-NN.csv};
                \addlegendentry{512}
                \addplot+
                        plot table[discard if not={VLEN}{1024}, x=matrix_size,y={vlen_improvement},col sep=comma] {./data/saturn_tests_vlen_int8_muRISCV-NN.csv};
                \addlegendentry{1024}
			\end{axis}
		\end{tikzpicture}
        \begin{tikzpicture}
			\begin{axis}[
                    title=Ours (int8),
                    ymin=-15,
                    ymax=15,
                    width=0.25\textwidth,
                    height=0.2\textheight,
                    ybar,
                    bar width=0.1cm,
                    enlarge y limits=false,
                    legend columns=2,
                    legend style={at={(0.01,0.98)},anchor=north west, font=\footnotesize},
                    legend cell align={left},
                    x label style={font=\footnotesize},
                    x tick label style={rotate=90, font=\footnotesize},
                    y tick label style={font=\footnotesize},
					ytick pos=left,
                    symbolic x coords={
                        16,
                        32,
                        64,
                        128,
                        256,
                        512,
                    },
                    xtick=data,
                    xticklabel=\empty,
				    cycle list name=ColorBlindFriendlyCycleListBar,
				]
                
                \addplot+
                        plot table[discard if not={VLEN}{512}, x=matrix_size,y={vlen_improvement},col sep=comma] {./data/saturn_tests_vlen_int8_tuned-RVV.csv};
                \addplot+
                        plot table[discard if not={VLEN}{1024}, x=matrix_size,y={vlen_improvement},col sep=comma] {./data/saturn_tests_vlen_int8_tuned-RVV.csv};
			\end{axis}
		\end{tikzpicture}

        \begin{tikzpicture}
			\begin{axis}[
                    title=Ours (float16),
                    ymin=-15,
                    ymax=15,
                    width=0.25\textwidth,
                    height=0.2\textheight,
                    xlabel={Matrices dimensions},
                    ylabel={Speedup [\%]},
                    ybar,
                    bar width=0.1cm,
                    enlarge y limits=false,
                    legend columns=2,
                    legend style={at={(0.01,0.98)},anchor=north west, font=\footnotesize},
                    legend cell align={left},
                    x label style={font=\footnotesize},
                    x tick label style={rotate=90, font=\footnotesize},
                    y tick label style={font=\footnotesize},
					ytick pos=left,
                    symbolic x coords={
                        16,
                        32,
                        64,
                        128,
                        256,
                        512,
                    },
                    xtick=data,
				    cycle list name=ColorBlindFriendlyCycleListBar,
				]
                
                \addplot+
                        plot table[discard if not={VLEN}{512}, x=matrix_size,y={vlen_improvement},col sep=comma] {./data/saturn_tests_vlen_float16_tuned-RVV.csv};
                \addplot+
                        plot table[discard if not={VLEN}{1024}, x=matrix_size,y={vlen_improvement},col sep=comma] {./data/saturn_tests_vlen_float16_tuned-RVV.csv};
			\end{axis}
		\end{tikzpicture}
        \hspace{0.1pt}
        \begin{tikzpicture}
			\begin{axis}[
                    title=Ours (float32),
                    ymin=-15,
                    ymax=15,
                    width=0.25\textwidth,
                    height=0.2\textheight,
                    xlabel={Matrices dimensions},
                    ybar,
                    bar width=0.1cm,
                    enlarge y limits=false,
                    legend columns=2,
                    legend style={at={(0.01,0.98)},anchor=north west, font=\footnotesize},
                    legend cell align={left},
                    x label style={font=\footnotesize},
                    x tick label style={rotate=90, font=\footnotesize},
                    y tick label style={font=\footnotesize},
					ytick pos=left,
                    symbolic x coords={
                        16,
                        32,
                        64,
                        128,
                        256,
                        512,
                    },
                    xtick=data,
				    cycle list name=ColorBlindFriendlyCycleListBar,
				]
                
                \addplot+
                        plot table[discard if not={VLEN}{512}, x=matrix_size,y={vlen_improvement},col sep=comma] {./data/saturn_tests_vlen_float32_tuned-RVV.csv};
                \addplot+
                        plot table[discard if not={VLEN}{1024}, x=matrix_size,y={vlen_improvement},col sep=comma] {./data/saturn_tests_vlen_float32_tuned-RVV.csv};
			\end{axis}
		\end{tikzpicture}
        \caption{Impact of VLEN on the execution time of matrix multiplications on the Saturn Vector Unit. The speedup baseline for each benchmark, datatype and target (muRISCV-NN or ours) is the execution time of the same matrix multiplication compiled with the same target but with VLEN = 256.}
        \label{fig:saturn_vlen}
    \end{figure}

    For the Saturn Vector Unit, as seen in Figure \ref{fig:saturn_1024_test}, enabling the autovectorization feature of GCC on the non-tuned C code does not necessarily mean an automatic improvement in the latency of the program. For the \textit{int8} versions, muRISCV-NN is able to accelerate the execution much more efficiently than the auto-vectorization feature from GCC. Nevertheless, our proposal surpasses all other for all experiments, including \textit{float16} and \textit{float32} versions, which muRISCV-NN does not support. In average, our proposal shows a mean improvement of 84\% with respect to the autovectorization feature of GCC 14, and of 50\% when compared against muRISCV-NN. For SoCs with VLEN = 256 and 512 we observe the same results.

    As explained, implementing these SoCs on an FPGA allows us to evaluate vector units with different VLEN. As such, in Figure \ref{fig:saturn_vlen} we vary VLEN from 256 to 1024 and compare the performance of muRISCV-NN against the tuned schedules found by our work. For muRISCV-NN, we see that naively increasing VLEN actually generates a negative impact on the performance of individual matrix multiplications. This gives even more weight to our initial claim that hand-crafted kernels are not suitable for different hardware configurations. On the other hand, the schedules found by our kernels mitigate this effect, as we tune each matrix multiplication for each hardware configuration. We also observe that, if the found candidates get worse when VLEN increases, the latency reduction is not that big as with muRISCV-NN.

    \begin{figure}[!t]
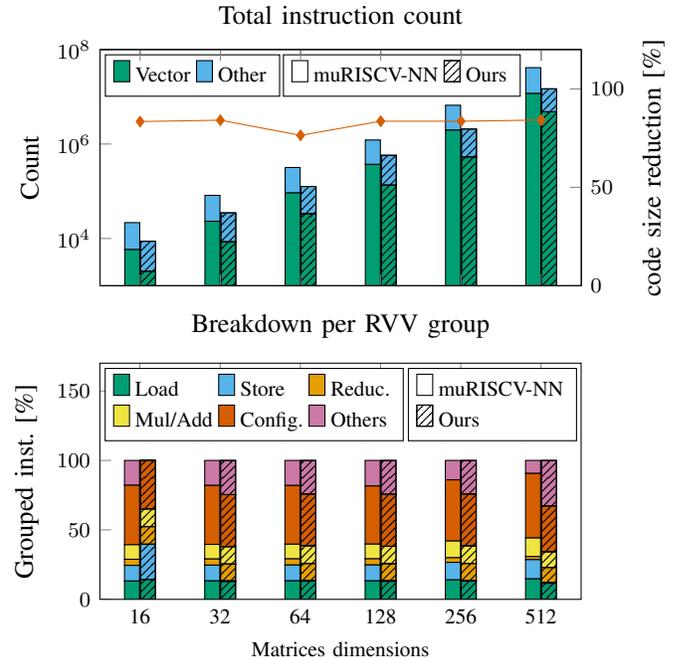


        \caption{Analysis of the instruction execution traces measured in QEMU for VLEN=1024 (line chart corresponds to the axis on the right).}
        \label{fig:qemu_instruction_traces}
    \end{figure}

    \begin{figure*}[!t]
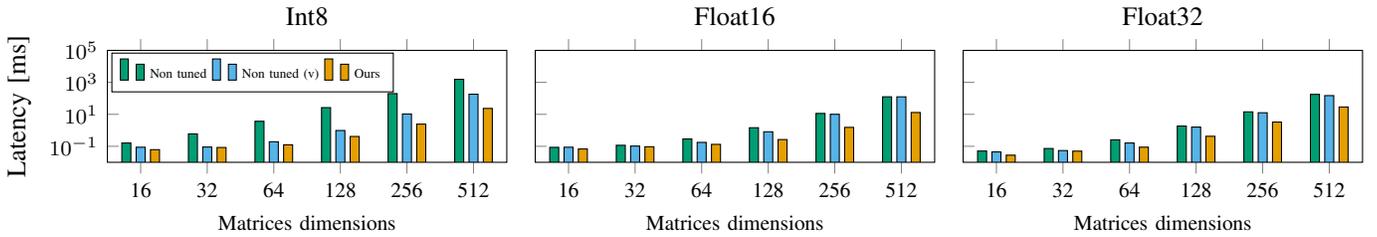

			%
        \hspace*{\fill}
        \caption{Benchmarking of matrix multiplications on the Banana Pi-F3 (VLEN=256)}
        \label{fig:bpi_test}
    \end{figure*}

    \begin{figure*}[!t]
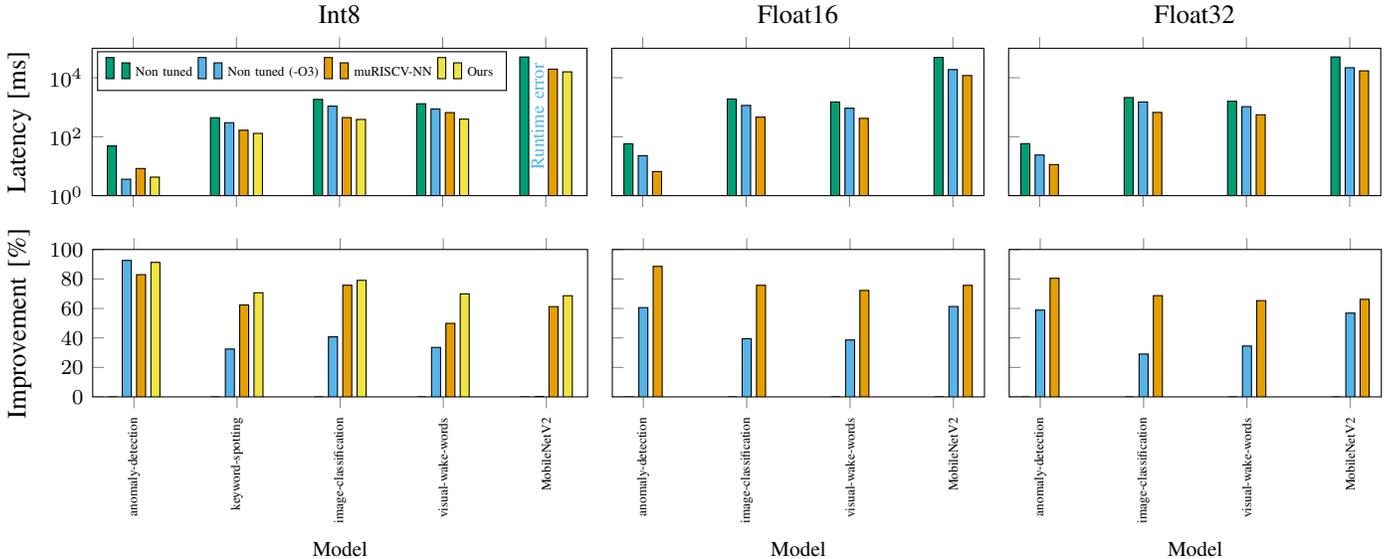

            %
        \caption{Complete models on the Saturn Vector Unit (VLEN=1024). Improvement is calculated using "Non tuned" as baseline.}
        \label{fig:models_saturn}
    \end{figure*}

    To corroborate why our schedules are better than the ones provided by muRISCV-NN, we record instruction traces during the execution of the model using a QEMU TCG plugin. We then grouped the executed vector instructions into groups (load, store, configuration, etc). Figure \ref{fig:qemu_instruction_traces} presents our results. In terms of absolute total instruction count, we are able to observe that our schedules use much fewer instructions than muRISCV-NN for all tests. The same happens if we compare only vector instructions. In reference to the total amount of vector instructions executed, although both execute a similar amount of relative \textit{Load} and \textit{Mult/Add} instructions, muRISCV-NN executes a significant amount of \textit{Store} instructions, which our schedules keep at a minimum (less than 1\% of the total executed vector instructions\footnote{Except for the matrix multiplication of size 16, because for VLEN = 1024, J = 32, so MetaSchedule selects the \textit{implementation} with J = 1, as explained in Section \ref{section:proposal}.}), thanks to the accumulation of output data provided by the \textit{vslideup} instruction in our intrinsic (Algorithm \ref{fig:rvv_multivmul}). Although we plot only the results for VLEN = 1024, we observe the same behavior for 512 and 256. As such, we are able to conclude that our schedules utilize the vector registers much more efficiently by executing more operations on the data already available in them before copying them out to memory.

    Additionally, Figure \ref{fig:qemu_instruction_traces} (top) also shows the size reduction of the matrix multiplication code in the final binary when using our proposal in comparison to muRISCV-NN. We observe that our proposal not only executes fewer instructions during the execution, but it also reduces the size of the code section in the final binary by around 90\%.

    For the matrix multiplications tuned on the Banana Pi board, Figure \ref{fig:bpi_test} shows that the kernels found by our proposal outperform the ones generated by LLVM with autovectorization enabled for all evaluated data types, with a mean 50\% improvement in execution latency. 

    \subsection{Complete networks}
    
    To demonstrate the performance of full networks accelerated using RVV instructions, we evaluate workloads from the MLPerf Tiny Benchmark \cite{DBLP:journals/corr/abs-2106-07597} (anomaly-detection, keyword-spotting, image-classification and visual-wake-words), standard Convolutional Neural Networks (CNN) like MobileNetv2 \cite{DBLP:journals/corr/abs-1801-04381} or ResNet18 \cite{he2015deepresiduallearningimage}, Transformer networks for the area of Natural Language Processing (NLP) like BERT \cite{devlin2019bertpretrainingdeepbidirectional}, and Generative Adversarial Networks like DCGAN \cite{radford2016unsupervisedrepresentationlearningdeep}. Additionaly, we also evaluate Largue Language Models (LLMs), like MobileLLM \cite{liu2024mobilellmoptimizingsubbillionparameter} (in its 125 million parameter version) \footnote{Although MobileLLM is an example of a sub-billion parameter LLM, it still requires significant amounts of memory for weights, so we only tune it and evaluate it on the Banana Pi board}. For MobileNetv2 and ResNet18, we used an input image size with dimensions $(224,224,3)$. For BERT, we deploy the tiny version of the network with a sequence length of 64 (the same sequence length used for MobileLLM). For DCGAN, the input latent space has a dimension of $(1,100)$.

    \begin{figure}[!t]
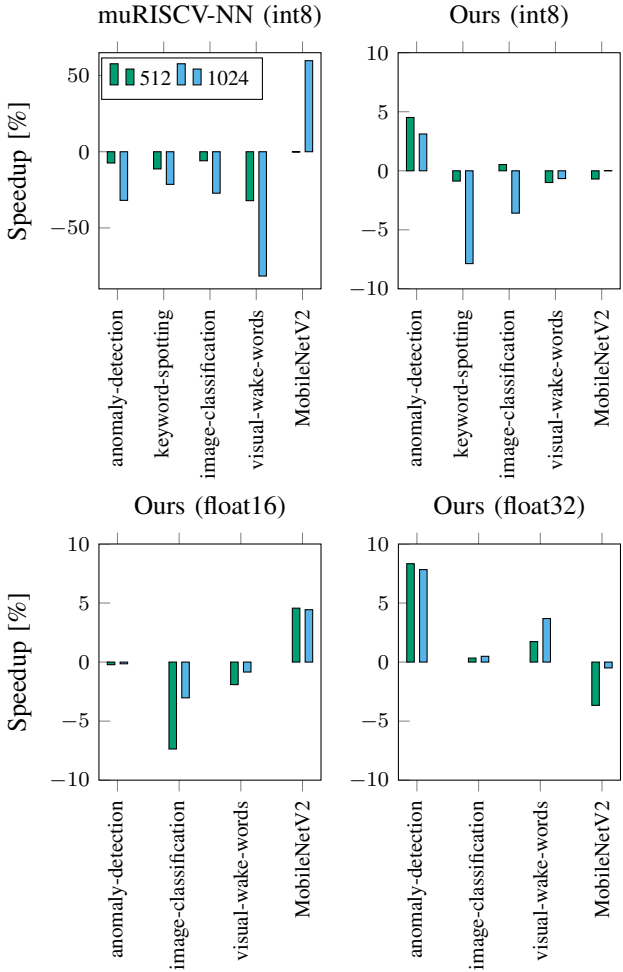


        \caption{Impact of VLEN on the execution time of complete networks on the Saturn Vector Unit. The speedup baseline for each model, datatype and target (muRISCV-NN or ours) is the execution time of the same model compiled with the same target but with VLEN = 256.}
        \label{fig:saturn_models_vlen}
    \end{figure}
    
    \begin{figure}[!t]
			%
        \caption{Analysis of the instruction execution traces measured in QEMU for VLEN=1024 (line chart corresponds to the axis on the right).}
        \label{fig:qemu_instruction_traces_models}
    \end{figure}  
    
    \begin{figure*}[!t]
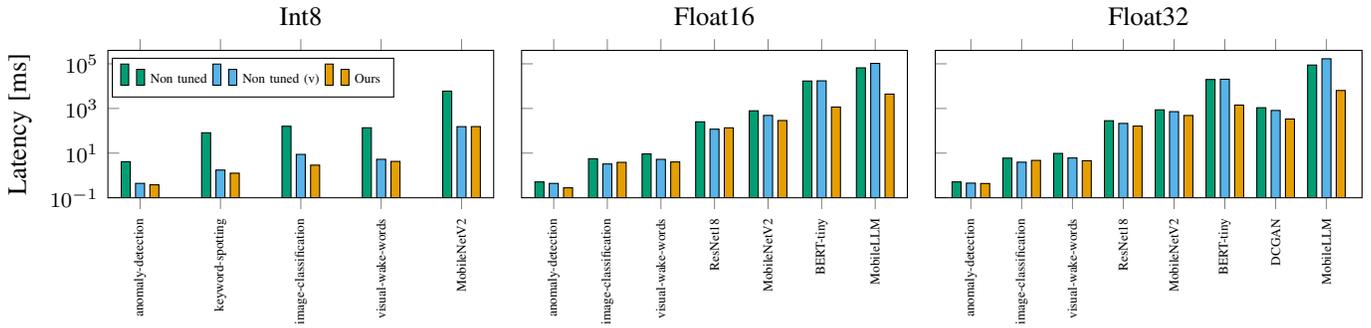

		\centering    
        %

        \caption{Complete models on the Banana Pi-F3 (VLEN=256).}
        \label{fig:models_bpi}
    \end{figure*}

    We configured MetaSchedule to evaluate a maximum of 200 tensor candidates per network (except for the LLM model, where, given the amount of layers in the model, we increased the limit to 400 in order to measure at least 10 schedule candidates for each layer). This keeps the search space process bound and finishes in a timely fashion.  
    
    For the models deployed on the Saturn Vector Unit (Figure \ref{fig:models_saturn}), we observe that the faster networks are found by our proposal, improving against both the auto-vectorization from GCC 14 (by 46\%) and even muRISCV-NN (by 29\% for the \textit{int8} models). We observe the same behavior for VLEN = 256 and 512. Only in one case (\textit{int8} anomaly-detection) was the auto-vectorization able to find networks with a similar performance to the ones found by our proposal, but this can probably be improved even more by forcing MetaSchedule to measure more candidates. When looking at the impact of VLEN on the execution of the complete models in Figure \ref{fig:saturn_models_vlen}, the same behavior seen in Section \ref{section:eval_matrix} can be appreciated. The same happens when looking at the instruction traces obtained with QEMU (Figure \ref{fig:qemu_instruction_traces_models}). We still see that, in terms of absolute instruction count, our proposal executes fewer instructions, although for complete networks, it seems our schedules execute more vector instructions. Nevertheless, we still observe that our schedules execute a fewer amount of relative \textit{Store} instructions.

    In terms of code size reduction (Figure \ref{fig:qemu_instruction_traces_models}, top), we observe a significant reduction of around 90\% for all models, except \textit{anomaly-detection}, where our code results much bigger than the one from muRISCV-NN. This happens because this network is comprised only of fully connected layers, which all call the same function of the muRISCV-NN library. Our proposal generates instead specific code for each layer because it finds different optimal schedules for each of them, thus the code size is bigger.
    
    For the evaluation on the Banana Pi-F3 board (Figure \ref{fig:models_bpi}), we observe that our proposal is able to accelerate networks almost always better than the autovectorization feature of LLVM, with a mean 35\% improvement in execution latency.  Even in the cases where our solutions are not better, they are still comparable, and probable more efficient solutions could be found by increasing the amount of candidates measured by MetaSchedule.

\section{Conclusion}
\label{section:conclusions}

    In this paper, we presented an extension of TVM's MetaSchedule framework to enable the tuning of AI workloads on RISC-V CPUs supporting the RVV 1.0 Vector Extension. We described the \textit{tensor intrinsics} integrated into TVM, and proceeded to tune a wide range of AI workloads, including tasks related to Computer Vision, Natural Language Processing, Adversarial Networks, and even Large Language Models. We implemented multiple SoCs with different vector unit parameters on an FPGA and evaluated on them the programs found by our integration against a previous work (muRISCV-NN) and the autovectorization feature of GCC 14. Our proposal is able to provide a 50\% and 84\% speedup respectively when tuning single matrix multiplications, and 29\% and 46\% when tuning complete AI models. We also analyzed the instruction traces of the execution of the models to justify why the schedules found by our proposal are faster than muRISCV-NN. We even showed that the code size of the compiled models is significantly smaller than the ones for the same models compiled with muRISCV-NN. Finally, we also used our integration to target a commercially available board, and found our proposal provides a 35\% speedup for complete AI models when compared against the standard LLVM 19 autovectorization.

    In terms of limitations, although in Section \ref{section:eval} we demonstrated the advantages of our proposal, the tuning process still requires some time to find efficient mappings. For cases where rapid prototyping of quantized AI workloads is required, relying on libraries of hand-crafted kernels like muRISCV-NN already provides good mappings. But the speedups measured in Section \ref{section:eval} strongly suggest that spending the time required to execute MetaSchedule with our proposal before deployment on the final product is a small price to pay for huge latency improvements.
    

    The open-source nature of this work will allow the community to extend it to target other RISC-V extensions. This is particularly interesting for embedded devices implementing more specific extensions, like the Packed SIMD extension. But HPC applications can also benefit from this integration, as the LLVM-based implementations of our tensor intrinsics enable the execution of networks using the full capabilities of the TVM runtime.

\balance

\bibliographystyle{IEEEtran}
\bibliography{bib}


\end{document}